\documentclass[11pt,a4paper]{article}
\usepackage[hyperref]{acl2020}
\usepackage{times}
\usepackage{latexsym}

\usepackage{microtype}
\usepackage{amsmath}
\usepackage{amsthm}
\usepackage{amsfonts}
\usepackage{amssymb}
\usepackage{graphicx}
\usepackage{booktabs}
\usepackage{placeins}
\usepackage{multirow}
\usepackage{tabularx}
\usepackage{epstopdf}
\usepackage{color}
\usepackage{subcaption}
\usepackage[T1]{fontenc}
\usepackage{amsmath,amsfonts,bm}
\usepackage{balance}
\usepackage{enumitem}
\usepackage{xcolor}
\usepackage{balance}
\usepackage{wrapfig}
\usepackage{enumitem}
\usepackage{graphics}
\usepackage{makecell}
\usepackage[utf8]{inputenc}
\usepackage[english]{babel}

\usepackage{soul}
\usepackage{float}

\aclfinalcopy 
\def\aclpaperid{1604} 

\title{
A Re-evaluation of Knowledge Graph Completion Methods
}

\author{Zhiqing Sun$^1$\thanks{\ \ Equal contribution.} \quad Shikhar Vashishth$^{1,2*}$ \quad Soumya Sanyal$^{2*}$  \\ \textbf{Partha Talukdar}$^{2}$ \quad 	\textbf{Yiming Yang}$^1$\\
	$^1$ Carnegie Mellon University, \
	$^2$ Indian Institute of Science\\
	{\tt \small \{zhiqings,svashish,yiming\}@cs.cmu.edu} \\
	{\tt \small \{soumyasanyal,ppt\}@iisc.ac.in} \\
}

\date{}

\begin{document}
\maketitle

 \newcommand{\refalg}[1]{Algorithm \ref{#1}}
\newcommand{\refeqn}[1]{Equation \ref{#1}}
\newcommand{\reffig}[1]{Figure \ref{#1}}
\newcommand{\reftbl}[1]{Table \ref{#1}}
\newcommand{\refsec}[1]{Section \ref{#1}}

\newcommand{\datafb}{FB15k-237}
\newcommand{\datawn}{WN18RR}
\newcommand{\datayago}{YAGO3-10}

\newcommand{\reminder}[1]{\textcolor{red}{[[ #1 ]]}\typeout{#1}}
\newcommand{\reminderR}[1]{\textcolor{gray}{[[ #1 ]]}\typeout{#1}}

\newcommand{\add}[1]{\textcolor{red}{#1}\typeout{#1}}
\newcommand{\remove}[1]{\sout{#1}\typeout{#1}}

\newcommand{\m}[1]{\mathcal{#1}}
\newcommand{\method}{SynGCN}
\newcommand{\methods}{WG}
\newcommand{\methodside}{SemGCN}
\newcommand{\methodsyn}{SynGCN}
\newcommand{\methodsidefull}{Semantic-GCN}
\newcommand{\methodsynfull}{Syntactic-GCN}

\newcommand{\problem}{DD}
\newcommand{\problemfull}{Document Dating}

\newcommand{\tensor}{\mathcal{X}}
\newcommand{\real}{\mathbb{R}}

\newcommand{\tuples}{\mathbb{T}}

\newcommand{\argmax}{arg\,max}

\newcommand\norm[1]{\left\lVert#1\right\rVert}

\newcommand{\note}[1]{\textcolor{blue}{#1}}

\newcommand*{\Scale}[2][4]{\scalebox{#1}{$#2$}}%
\newcommand*{\Resize}[2]{\resizebox{#1}{!}{$#2$}}%
\definecolor{officegreen}{rgb}{0.0, 0.5, 0.0}
\def\mat#1{\mbox{\bf #1}}

\theoremstyle{definition}
\newtheorem{definition}{Definition}[section]

\newtheorem{proposition}{Proposition}[section]
\newtheorem*{lemma*}{Lemma}

\begin{abstract}
Knowledge Graph Completion (KGC) aims at automatically predicting missing links for large-scale knowledge graphs. A vast number of state-of-the-art KGC techniques have got published at top conferences in several research fields, including data mining, machine learning, and natural language processing. However, we notice that several recent papers report very high performance, which largely outperforms previous state-of-the-art methods. 
In this paper, we find that this can be attributed to the inappropriate evaluation protocol used by them and propose a simple evaluation protocol to address this problem. The proposed protocol is robust to handle bias in the model, which can substantially affect the final results.
We conduct extensive experiments and report performance of several existing methods using our protocol. The reproducible code has been made publicly available.
\end{abstract}

\section{Introduction}
\label{sec:intro}

Real-world knowledge bases are usually expressed as multi-relational graphs, which are collections of factual triplets, where each triplet $(h, r, t)$ represents a relation $r$ between a head entity $h$ and a tail entity $t$.
However, real-word knowledge bases are usually incomplete \cite{kg_incomp1}, which motivates the research of automatically predicting missing links.
A popular approach for Knowledge Graph Completion (KGC) is to embed entities and relations into continuous vector or matrix space, and use a well-designed score function $f(h, r, t)$ to measure the plausibility of the triplet $(h, r, t)$.
Most of the previous methods use \textit{translation distance} based \cite{transe,transh,transg,rotate} and \textit{semantic matching} based \cite{rescal2013,distmult,hole,complex,analogy} scoring functions which are easy to analyze. 

However, recently, a vast number of neural network-based methods have been proposed. They have complex score functions which utilize black-box neural networks including Convolutional Neural Networks (CNNs) \cite{conve,convkb}, Recurrent Neural Networks (RNNs) \cite{ptranse,dolores}, Graph Neural Networks (GNNs) \cite{r_gcn,sacn_paper}, and Capsule Networks \cite{capse}. While some of them report state-of-the-art performance on several benchmark datasets that are competitive to previous embedding-based approaches, a considerable portion of recent neural network-based papers report very high performance gains which are not consistent across different datasets. Moreover, most of these unusual behaviors are not at all analyzed. Such a pattern has become prominent and is misleading the whole community.

%
%

In this paper, we investigate this problem and find that this is attributed to the inappropriate evaluation protocol used by these approaches. 
We demonstrate that their evaluation protocol gives a perfect score to a model that always outputs a constant irrespective of the input. This has lead to artificial inflation of performance of several models. 
For this, we find a simple evaluation protocol that creates a fair comparison environment for all types of score functions. We conduct extensive experiments to re-examine some recent methods and fairly compare them with existing approaches. The source code of the paper has been publicly available at \url{http://github.com/svjan5/kg-reeval}.


\section{Background}

\paragraph{Knowledge Graph Completion} Given a Knowledge Graph $\m{G} = (\m{E}, \m{R},\m{T})$, where $\m{E}$ and $\m{R}$ denote the set of entities and relations and $\m{T} = \{(h,r,t) \ | \ h,t \in \m{E}, r \in \m{R} \}$ is the set of triplets (facts), the task of Knowledge Graph Completion (KGC) involves inferring missing facts based on the known facts. Most the existing methods define an embedding for each entity and relation in $\m{G}$, i.e., $\bm{e}_h, \bm{e}_r \ \forall h \in \m{E}, r \in \m{R}$ and a score function $f(h,r,t): \m{E} \times \m{R} \times \m{E} \rightarrow \real{}$ which assigns a high score for valid triplets than the invalid ones.

\paragraph{KGC Evaluation} During KGC evaluation, for predicting $t$ in a given triplet $(h,r,t)$, a KGC model scores all the triplets in the set $\m{T}' = \{(h,r,t') \ | \ t' \in \m{E}\}$. Based on the score, the model first sorts all the triplets and subsequently finds the rank of the valid triplet $(h,r,t)$ in the list. In a more relaxed setting called \textit{filtered setting}, all the known correct triplets (from train, valid, and test triplets) are removed from $\m{T}'$ except the one being evaluated \cite{transe}.
The triplets in $\m{T}'  - \{t\}$ are called negative samples.

\paragraph{Related Work}
\label{sec:related_works}

Prior to our work, \citet{baselines_strike_back,mausam_2} cast doubt on the claim that performance improvement of several models is due to architectural changes as opposed to hyperparameter tuning or different training objective. 
In our work, we raise similar concerns but through a different angle by highlighting issues with the evaluation procedure used by several recent methods. \citet{kg_geometry} analyze the geometry of KG embeddings and its correlation with task performance while \citet{effect_of_loss_function} examine the effect of different loss functions on performance. Also, \citet{mausam_1} investigate evaluation protocols for handling out-of-vocabulary entities. However, their analysis is restricted to non-neural approaches.


\begin{table}[t]
	\centering
	\resizebox{0.9\linewidth}{!}{
	\begin{tabular}{lll}
		\toprule
		& \multicolumn{1}{l}{\textbf{\datafb{}}} & \multicolumn{1}{l}{\textbf{\datawn{}}}\\ 
		\midrule
		ConvE & .325  & .430 \\
		\midrule 
		RotatE & .338 (+4.0\%)& .476 (+10.6\%)\\
		TuckER &  .358 (+10.2\%)& .470 (+9.3\%)\\
		\midrule
		ConvKB & .396 (+21.8\%) & .248 (-42.3\%)\\
		CapsE    & .523 (+60.9\%) & .415 (-3.4\%)\\
		KBAT       & .518 (+59.4\%) & .440 (+2.3\%)\\
		TransGate & .404  (+24.3\%) & .409  (-4.9\%)\\
		
		\bottomrule
		\addlinespace
	\end{tabular}
	}
	\caption{\label{tbl:dataset_comp}Changes in MRR for different methods on \datafb{} and \datawn{} datasets with respect to ConvE show inconsistent improvements. 
	}
\end{table}
\section{Observations}

In this section, we first describe our observations and concerns and then investigate the reason behind.

\subsection{Inconsistent Improvements over Benchmark Datasets}
\label{sec:results_datasets}
Several recently proposed methods report high performance gains on a particular dataset. However, their performance on another dataset is not consistently improved.
In Table \ref{tbl:dataset_comp}, we report change in MRR score on \datafb{} \cite{toutanova} and \datawn{} \cite{conve} datasets with respect to ConvE \cite{conve} for different methods including RotatE \cite{rotate}, TuckER \cite{tucker}, ConvKB \cite{convkb}, CapsE \cite{capse}, KBAT \cite{kbat}, and TransGate \cite{transgate}. Overall, we find that for a few recent NN based methods, there are inconsistent gains on these two datasets. For instance, in ConvKB, there is a 21.8\% improvement over ConvE on \datafb{}, but a degradation of 42.3\% on \datawn{}, which is surprising given the method is claimed to be better than ConvE. On the other hand, methods like RotatE and TuckER give consistent improvement across both benchmark datasets. 

\begin{figure}[t]
	\centering
	\includegraphics[width=\columnwidth]{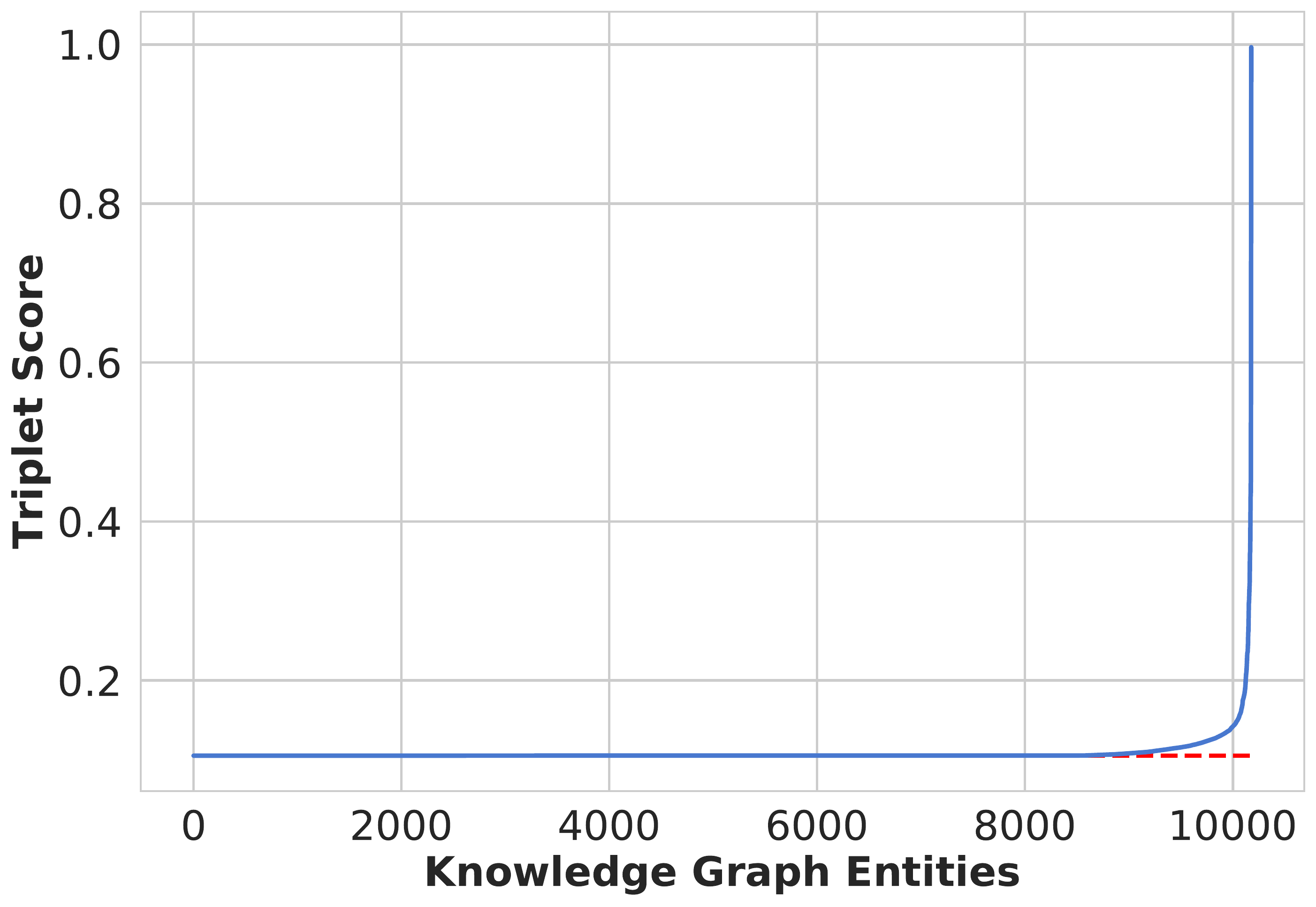}
	\caption{\label{fig:score_dist} Sorted score distribution of ConvKB for an example valid triplet and its negative samples. The score is normalized into $[0, 1]$ (lower the better). Dotted line indicate the score for the valid triplet. We find that in this example, around 58.5\% negative sampled triplets obtain the exact same score as the valid triplet. 
	}  
\end{figure}

\subsection{Observations on Score Functions}
\label{sec:score_dist}

\paragraph{Score distribution}
When evaluating KGC methods, for a given triplet $(h,r,t)$, the ranking of $t$ given $h$ and $r$ is computed by scoring all the triplets of form $\{(h,r,t') \ |\  t' \in \m{E}\}$, where $\m{E}$ is the set of all entities. On investing a few recent NN based approaches, we find that they have unusual score distribution, where some negatively sampled triplets have the same score as the valid triplet. An instance of \datafb{} dataset is presented in Figure \ref{fig:score_dist}. Here, out of 14,541 negatively sampled triplets, 8,520 have the exact same score as the valid triplet.


\begin{figure}[t]
	\centering
	\includegraphics[width=0.87\columnwidth]{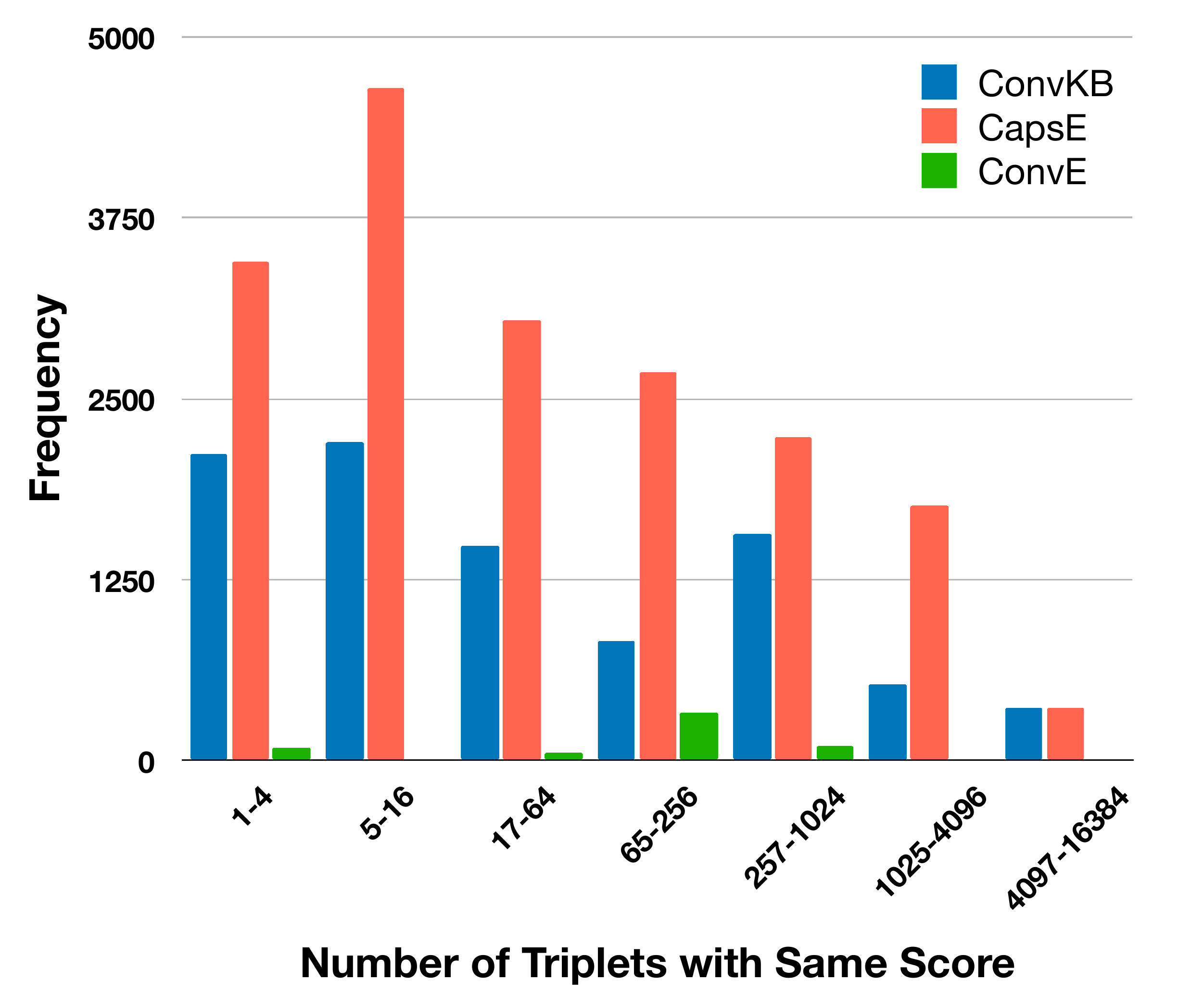}
	\caption{\label{fig:samescr_plot}Plot shows the frequency of the number of negative triplets with the same assigned score as the valid triplet during evaluation on \datafb{} dataset. The results show that for methods like ConvKB and CapsE, a large number of negative triplets get the same score as the valid triplets whereas for methods like ConvE such occurrences are rare.
	} 
\end{figure}

\paragraph{Statistics on the whole dataset}
In Figure \ref{fig:samescr_plot}, we report the total number of triplets with the exact same score over the entire dataset for ConvKB \cite{convkb} and CapsE \cite{capse} and compare them with ConvE \cite{conve} which does not suffer from this issue. We find that both ConvKB and CapsE have multiple occurrences of such unusual score distribution. On average, ConvKB and CapsE have 125 and 197 entities with exactly same score as the valid triplet over the entire evaluation dataset of \datafb{}, whereas ConvE has around 0.002, which is almost negligible. In Section \ref{sec:details}, we demonstrate how this leads to massive performance gain for methods like ConvKB and CapsE. 

\begin{figure}[t]
	\centering
	\includegraphics[width=\columnwidth]{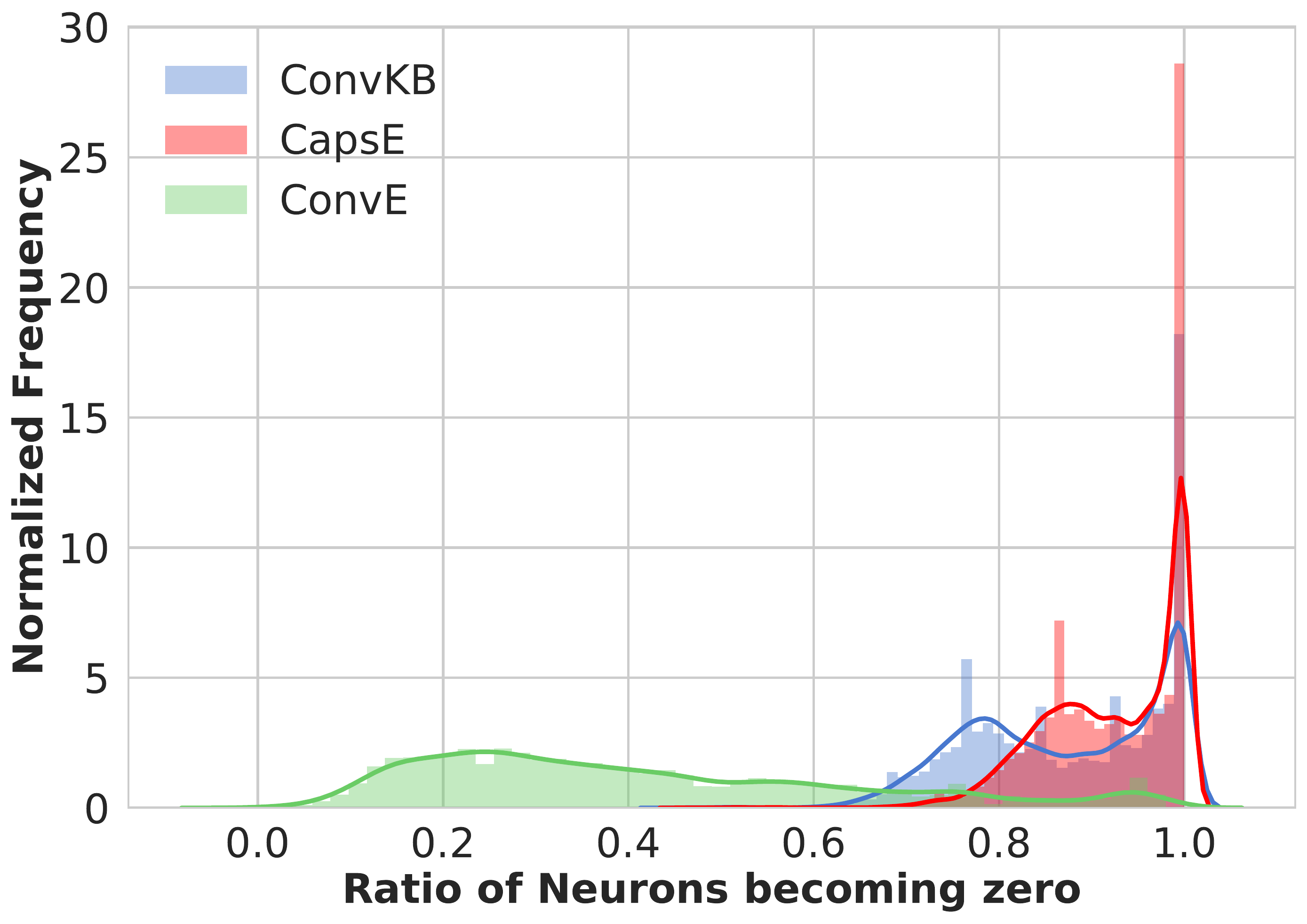}
	\caption{\label{fig:neuron_plot}Distribution of ratio of neurons becoming zero after ReLU activation in different methods for the valid triplets in \datafb{} dataset. We find that for ConvKB and CapsE an unusually large fraction of neurons become zero after ReLU activation whereas the does not hold with ConvE. 
	}
\end{figure}

\paragraph{Root of the problem}
Further, we investigate the cause behind such unusual score distribution. In Figure \ref{fig:neuron_plot}, we plot the ratio of neurons becoming zero after ReLU activation for the valid triplets vs. their normalized frequency on \datafb{} dataset. The results show that in ConvKB and CapsE, a large fraction (87.3\% and 92.2\% respectively) of the neurons become zeros after applying ReLU activation. However, with ConvE, this count is substantially less (around 41.1\%). Because of the zeroing of nearly all neurons (at least 14.2\% for ConvKB and 22.0\% for CapsE), the representation of several triplets become very similar during forward pass and thus leading to obtaining the exact same score. 


\section{Evaluation Protocols for KGC}
\label{sec:details}

In this section, we present different evaluation protocols that can be adopted 
in knowledge graph completion. We further show that inappropriate evaluation protocol is the key reason behind the unusual behavior of some recent NN-based methods.

\paragraph{How to deal with the same scores?}
An essential aspect of the evaluation method is to decide how to break ties for triplets with the same score. More concretely, while scoring the candidate set $\m{T}'$, if there are multiple triplets with the same score from the model, one should decide which triplet to pick. Assuming that the triplets are sorted in a stable manner, we design a general evaluation scheme for KGC, which consists of the following three different protocols:

\begin{itemize}
[itemsep=2pt,parsep=1pt,partopsep=1pt,leftmargin=*,topsep=1pt]
	\item \textbf{\textsc{Top}}: In this setting, the correct triplet is inserted in the beginning of $\m{T}'$.
	\item \textbf{\textsc{Bottom}}: Here, the correct triplet is inserted at the end of $\m{T}'$.
	\item \textbf{\textsc{Random}}: In this, the correct triplet is placed randomly in $\m{T}'$.
\end{itemize}

\setlength{\tabcolsep}{3.0pt}

\begin{table*}[t]
	\small
	\centering
	\begin{tabular}{lccc|ccc|ccc|ccc}
		\toprule
		& \multicolumn{3}{c}{\textbf{Reported}} & \multicolumn{3}{c}{\textbf{\textsc{Random}}} & \multicolumn{3}{c}{\textbf{\textsc{Top}}} & \multicolumn{3}{c}{\textbf{\textsc{Bottom}}} \\ 
		\cmidrule(r){2-4}  \cmidrule(r){5-7} \cmidrule(r){8-10} \cmidrule(r){11-13}
		& MRR $\uparrow$ & MR $\downarrow$ & H@10 $\uparrow$ & MRR  $\uparrow$ & MR $\downarrow$ & H@10 $\uparrow$ & MRR $\uparrow$ & MR $\downarrow$ & H@10 $\uparrow$ & MRR $\uparrow$ & MR $\downarrow$ & H@10 $\uparrow$ \\
		\midrule
		ConvE		& .325 & 244 & .501 	& .324 $\pm$ .0 & 285	$\pm$ 0 & .501 $\pm$ .0& .324	& 285 & .501	& .324 &	285	& .501 \\
		RotatE		& .338 & 177 & .533 	& .336 $\pm$ .0 & 178 $\pm$ 0 & .530 $\pm$ .0	& .336 & 178 & .530& .336 & 178 & .530 \\ 
		TuckER		& .358 & -   & .544 	& .353 $\pm$ .0	& 162 $\pm$ 0	& .536 $\pm$ .0	& .353	& 162	& .536 & .353	& 162	& .536 \\
		\midrule
		\midrule
		\multirow{2}{*}{ConvKB}		& \multirow{2}{*}{.396} & \multirow{2}{*}{257} & \multirow{2}{*}{.517} 	& \multirow{2}{*}{.243 $\pm$ .0}	& \multirow{2}{*}{309 $\pm$ 2}	& \multirow{2}{*}{.421 $\pm$ .0} & .407	& 246	& .527 & .130	& 373	& .383 \\
		& & & & & & & (+.164) & (-63)& (+.106) & (-.113) & (+64) & (-.038)\\
		\midrule
		\multirow{2}{*}{CapsE}		& \multirow{2}{*}{.523} & \multirow{2}{*}{303} & \multirow{2}{*}{.593} 	& \multirow{2}{*}{.150 $\pm$ .0} & \multirow{2}{*}{403 $\pm$ 2} & \multirow{2}{*}{.356 $\pm$ .0} & .511& 	305	& .586 & .134	& 502	& .297 \\
		& & & & & & & (+.361) & (-99) & (+.229) & (-.016) & (+99) & (-.059)\\
		\midrule
		KBAT		& .518\dag & 210\dag & .626\dag 	& .157 $\pm$ .0&	270	$\pm$ 0& .331 $\pm$ .0	& .157 &	270	& .331  & .157 &	270	& .331 \\
		\bottomrule
		\addlinespace
	\end{tabular}
	\caption{\label{tbl:fb15k_results}Effect of different evaluation protocols on recent KG embedding methods on \datafb{} dataset. 
	For \textsc{Top} and \textsc{Bottom}, we report changes in performance with respect to \textsc{Random} protocol. Please refer to Section \ref{sec:results_evaluation} for details. \dag: KBAT has test data leakage in their original implementation, which is fixed in our experiments.}
\end{table*}

\setlength{\tabcolsep}{6pt}




\paragraph{Discussion} Based on the definition of the three evaluation protocols, it is clear that \textsc{Top} evaluation protocol does not evaluate the model rigorously. It gives the models that have a bias to provide the same score for different triplets, an inappropriate advantage.
On the other hand, \textsc{Bottom} evaluation protocol can be unfair to the model during inference time because it penalizes the model for giving the same score to multiple triplets, i.e., if many triplets have the same score as the correct triple, the correct triplet gets the least rank possible.

As a result, \textsc{Random} is the best evaluation technique which is both rigorous and fair to the model.
It is in line with the situation we meet in the real world: given several same scored candidates, the only option is to select one of them randomly.
Hence, we propose to use \textsc{Random} evaluation scheme for all model performance comparisons.

\section{Experiments}
\label{sec:experimental_setup}

In this section, we conduct extensive experiments using our proposed evaluation protocols and make a fair comparison for several existing methods.

\subsection{Datasets}
\label{sec:datasets}
We evaluate the proposed protocols on \datafb{} \cite{toutanova} dataset\footnote{We also report our results on WN18RR \cite{conve} dataset in the appendix.}, which is a subset of FB15k \cite{transe} with inverse relations deleted to prevent direct inference of test triples from training.

\subsection{Methods Analyzed}
\label{sec:baselines}

In our experiments, we categorize existing KGC methods into the following two categories:

\begin{itemize}[itemsep=2pt,parsep=2pt,partopsep=0pt,leftmargin=*,topsep=4pt]
	\item \textbf{Non-Affected:} This includes methods which give consistent performance under different evaluation protocols. For experiments in this paper, we consider three such methods --
	ConvE, RotatE, and TuckER.
	\item \textbf{Affected:} This category consists of recently proposed neural-network based methods whose performance is affected by different evaluation protocols. ConvKB, CapsE, TransGate\footnote{Since we cannot find any open-source implementation of TransGate, we leave the re-evaluation of TransGate as our future work.}, and KBAT are methods in this category.
\end{itemize}

%
%
%

\subsection{Evaluation Metrics}
\label{sec:evaluation_metrics}

For all the methods, we use the code and the hyperparameters provided by the authors in their respective papers. 
Model performance is evaluated by Mean Reciprocal Rank (MRR), Mean Rank (MR) and Hits@10 (H@10) on the filtered setting \cite{transe}.

\subsection{Evaluation Results}
\label{sec:results}


%

\label{sec:results_evaluation}
To analyze the effect of different evaluation protocols described in Section \ref{sec:details}, we study the performance variation of the models listed in Section \ref{sec:baselines}. We study the effect of using \textsc{Top} and \textsc{Bottom} protocols and compare them to \textsc{Random} protocol. In their original paper, ConvE, RotatE, and TuckER use a strategy similar to the proposed \textsc{Random} protocol, while ConvKB, CapsE, and KBAT use \textsc{Top} protocol. We also study the random error in \textsc{Random} protocol with multiple runs, where we report the average and standard deviation on 5 runs with different random seeds. The results are presented in Tables \ref{tbl:fb15k_results}.

We observe that for Non-Affected methods like ConvE, RotatE, and TuckER, the performance remains consistent across different evaluation protocols. However, with Affected methods, there is a considerable variation in performance. Specifically, we can observe that these models perform best when evaluated using \textsc{Top} and worst when evaluated using \textsc{Bottom}\footnote{KBAT incorporates ConvKB in the last layer of its model architecture, which should be affected by different evaluation protocols. But we find another bug on the leakage of test triples during negative sampling in the reported model, which results in more significant performance degradation.}. Finally, we find that the proposed \textsc{Random} protocol is very robust to different random seeds. Although the theoretic upper and lower bounds of a \textsc{Random} score are \textsc{Top} and \textsc{Bottom} scores respectively, when we evaluate knowledge graph completion for real-world large-scale knowledge graphs, the randomness doesn't affect the evaluation results much. 



\section{Conclusion}
\label{sec:conclusion}
In this paper, we performed an extensive re-examination study of recent neural network based KGC techniques. We find that many such models have issues with their score functions. Combined with inappropriate evaluation protocol, such methods reported inflated performance. Based on our observations, we propose \textsc{Random} evaluation protocol that can clearly distinguish between these affected methods from others. We also strongly encourage the research community to follow the \textsc{Random} evaluation protocol for all KGC evaluation purposes.

\section*{Acknowledgements}
We thank the reviewers for their helpful comments. This work is supported in part by the National Science Foundation (NSF) under grant IIS-1546329 and Google PhD Fellowship.

\bibliography{acl2020}
\bibliographystyle{acl_natbib}

\appendix

\begin{center}
{\bf \large{
    Appendix}}
\end{center}

\section{Results on WN18RR dataset}
\begin{figure}[!htb]
	\centering
	\includegraphics[width=\columnwidth]{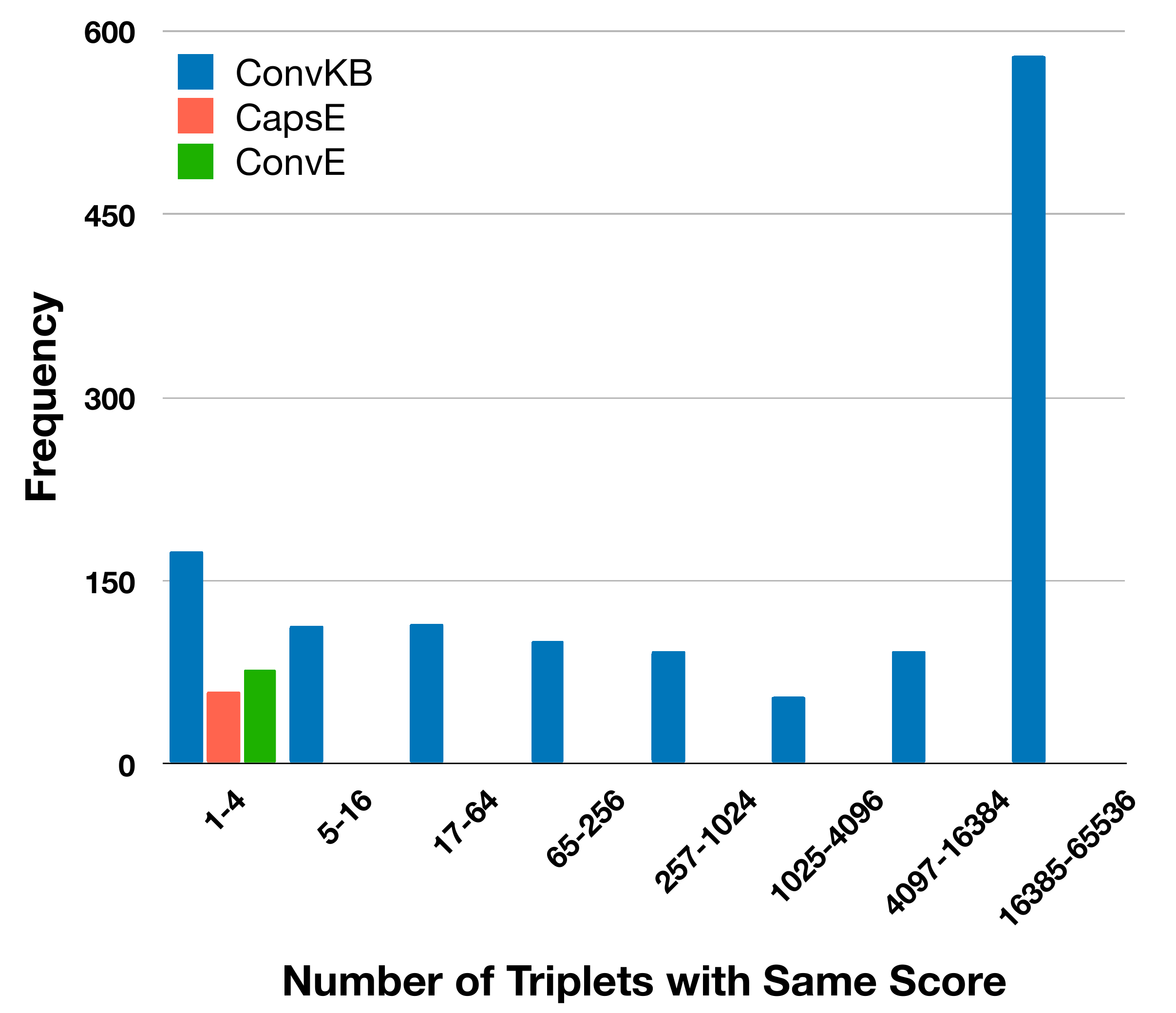}
	\caption{\label{fig:samescr_plot2}Plot shows the frequency of the number of negative triplets with the same assigned score as the valid triplet during evaluation on \datawn{} dataset. The results show that Unlike \datafb{}, in this dataset, only ConvKB has a large number of negative triplets get the same score as the valid triplets.
	} 
\end{figure}

Besides FB15k-237, we also evaluate the proposed protocols on \datawn{} \cite{conve} dataset, which is a subset of WN18 \cite{transe} containing lexical relations between words. Similar to \datafb{}, inverse relations are removed in \datawn{}. The results on WN18RR are shown in Table \ref{tbl:wn18_results}. From these results, we can draw similar conclusions as in Section 5. We also show the total number of triplets with the exact same score over the entire \datawn{} dataset for ConvKB, CapsE and ConvE in Figure \ref{fig:samescr_plot2}.

\setlength{\tabcolsep}{1.5pt}

\begin{table*}[!htb]
	\small
	\centering
	\begin{tabular}{lccc|ccc|ccc|ccc}
		\toprule
		&  \multicolumn{3}{c}{\textbf{Reported}} & \multicolumn{3}{c}{\textbf{\textsc{Random}}} & \multicolumn{3}{c}{\textbf{\textsc{Top}}} & \multicolumn{3}{c}{\textbf{\textsc{Bottom}}} \\ 
		\cmidrule(r){2-4}  \cmidrule(r){5-7} \cmidrule(r){8-10} \cmidrule(r){11-13}
		& MRR $\uparrow$ & MR $\downarrow$ & H@10 $\uparrow$ & MRR  $\uparrow$ & MR $\downarrow$ & H@10 $\uparrow$ & MRR $\uparrow$ & MR $\downarrow$ & H@10 $\uparrow$ & MRR $\uparrow$ & MR $\downarrow$ & H@10 $\uparrow$ \\
		\midrule
		ConvE 		& .43  & 4187 & .52  	& .444 $\pm$ .0	& 4950 $\pm$ 0	& .503 $\pm$ .0 	& .444	& 4950	& .503 	& .444	& 4950	& .503 \\
		RotatE 		& .476 &	3340 & .571 & .473 $\pm$ .0	& 3343 $\pm$ 0 & .571 $\pm$ .0 & .473 & 3343 & .571 & .473 & 3343 & .571 \\ 
		TuckER 		& .470 & -    & .526 	& .461 $\pm$ .0	 & 6324 $\pm$ 0	& .516 $\pm$ .0	& .461	 & 6324	& .516	& .461	 & 6324	& .516 \\
		\midrule
		\midrule
		\multirow{2}{*}{ConvKB} 		& \multirow{2}{*}{.248} & \multirow{2}{*}{2554} & \multirow{2}{*}{.525} 	& \multirow{2}{*}{.249 $\pm$ .0}	& \multirow{2}{*}{3433 $\pm$ 42}	&\multirow{2}{*}{.524 $\pm$ .0} & .251&	1696&	.529 & .164&	5168&	.516 \\
		& & & & & & & (+.002) & (-1737)& (+.005) & (-.085) & (+1735) & (-.008)\\
		\midrule
		\multirow{2}{*}{CapsE\ddag} 		& \multirow{2}{*}{.415} &  \multirow{2}{*}{719} & \multirow{2}{*}{.560} &
		\multirow{2}{*}{.415 $\pm$ .0}& \multirow{2}{*}{718 $\pm$ 0}	&\multirow{2}{*}{.559 $\pm$ .0} & \multirow{2}{*}{.415} &	\multirow{2}{*}{718}	& \multirow{2}{*}{.559} & .323	& 719 & .555 \\ 
		& & & & & & & & & & (-.092) & (+1) & (-.004)\\
		\midrule
		KBAT		& .440\dag  & 1940\dag  & .581\dag  	& .412 $\pm$ .0	& 1921 $\pm$ 0	& .554 $\pm$ .0 & .412	& 1921	& .554 & .412	& 1921	& .554\\
		\bottomrule
		\addlinespace
	\end{tabular}
	\caption{\label{tbl:wn18_results}Performance comparison under different evaluation protocols on \datawn{} dataset. For \textsc{Top} and \textsc{Bottom}, we report changes in performance with respect to \textsc{Random} protocol. \ddag: CapsE uses the pre-trained 100-dimensional Glove \cite{pennington2014glove} word embeddings for initialization on WN18RR dataset, which makes the comparison on WN18RR still unfair. \dag: KBAT has test data leakage in their original implementation, which is fixed in our experiments.}
\end{table*}

\setlength{\tabcolsep}{6pt}

\end{document}